\documentclass{article}
\usepackage{multirow}
\usepackage{PRIMEarxiv}

\usepackage{array}
\usepackage{makecell}
\usepackage{graphicx}
\usepackage{ucs} 
\usepackage[utf8x]{inputenc} 
\SetUnicodeOption{default} 
\usepackage[T1]{fontenc}    
\usepackage{hyperref}       
\usepackage{url}            
\usepackage{booktabs}       
\usepackage{amsfonts}       
\usepackage{nicefrac}       

\usepackage{fonts-tlwg}
\usepackage[main=english, thai]{babel}

\usepackage{microtype}      
\usepackage{lipsum}
\usepackage{fancyhdr}       
\usepackage{graphicx}       
\usepackage{framed}
\graphicspath{{media/}}     

\usepackage[symbol]{footmisc}
\title{WangchanLion and WangchanX MRC Eval}

\author{
\textbf{Wannaphong~Phatthiyaphaibun }\textsuperscript{\dag}\footnotemark[1] , \textbf{Surapon~Nonesung}\textsuperscript{\dag}\footnotemark[1] , \textbf{Patomporn~Payoungkhamdee}\textsuperscript{\dag},\textbf{Peerat~Limkonchotiwat}\textsuperscript{\dag},\\  \textbf{Can~Udomcharoenchaikit}\textsuperscript{\dag}, \textbf{Jitkapat~Sawatphol}\textsuperscript{\dag},
\textbf{Chompakorn Chaksangchaichot}\textsuperscript{\S},\\
\textbf{Ekapol~Chuangsuwanich}\textsuperscript{\S}, \textbf{Sarana~Nutanong}\textsuperscript{\dag}\\
  \textsuperscript{\dag}School of Information Science and Technology, VISTEC, Thailand\\
  \textsuperscript{\S}Department of Computer Engineering, Faculty of Engineering, Chulalongkorn University, Thailand \\
}

\begin{document}
\maketitle

\begin{abstract}
This technical report describes the development of WangchanLion, an instruction fine-tuned model focusing on Machine Reading Comprehension (MRC) in the Thai language. 
Our model is based on SEA-LION and a collection of instruction following datasets. 
To promote open research and reproducibility, we publicly release all training data, code, and the final model weights under the Apache-2 license. 
To assess the contextual understanding capability, we conducted extensive experimental studies using two Thai MRC datasets, XQuAD and Iapp\_wiki\_qa\_squad. 
Experimental results demonstrate the model's ability to comprehend the context and produce an answer faithful to the reference one in 0-shot and 1-shot settings.
In addition, our evaluation goes beyond the traditional MRC.
We propose a new evaluation scheme assessing the answer's correctness, helpfulness, conciseness, and contextuality.
%
%
Our code is available publicly at \url{https://github.com/vistec-AI/WangchanLion}.
\end{abstract}



\footnotetext[1]{Equal contribution: Wannaphong is the main evaluation contribution, Surapon is the main model contribution.\\
corresponding author: \{wannaphong.p\_s21,surapon.n\}@vistec.ac.th}

\section{Introduction}


Large Language Models (LLMs) have gained significant attention in the field of artificial intelligence in recent years due to their ability to generate human-like text and assist humans in various tasks. 
The applications of LLMs are vast and extend beyond just language generation, including language translation, question answering, and speech recognition. 
With the advent of advanced LLMs such as GPT-4 and ChatGPT, there has been an increasing interest in improving the capabilities of these models.

In addition to the proprietary API services, we see a strong interest in open-source research and development. 
For example, LLaMA 2~\cite{touvron2023llama} is a pretrained generative text model from Meta.
Mistral~\cite{jiang2023mistral} is a pretrained generative text model offered by Mistral AI. PolyLM~\cite{wei2023polylm} stands out as an open-source multilingual LLM developed by DAMO Academy, Alibaba Group. Additionally, SEA-LION\cite{sea_lion_2023} stands as a pretrained generative text model specific to Southeast Asia languages, developed by AI Singapore.
These open-source LLMs provide researchers and developers with greater flexibility to adjust the model's behavior through fine-tuning with their algorithms and datasets. In addition, there are models that support the Thai language but are not open-source that we consider to have good performance, such as SeaLLM (multilingual LLM for Southeast Asian)~\cite{nguyen2023seallms} and Typhoon-instruct (EN-TH instruction model by SCB10X)~\cite{pipatanakul2023typhoon}, which we will use the evaluation results to compare.

While these models support multiple languages, the models' capability on medium-to-low-resource languages like Thai and many other Southeast Asian languages is still greatly overlooked.
One issue that hinders the adoption of open-source LLMs in such languages is the ability to transfer knowledge from rich-resource languages due to the limited vocabulary size. 
To illustrate, LLaMA2's training dataset comprises approximately 2.0 trillion tokens, with over 89\% of them being in English. 
Vocabulary tokens in LLaMA2 are subwords rather than full words, as the LLaMA2 tokenizer employs Byte-Pair Encoding (BPE). 
Consequently, tokens in LLaMA2 are not inherently tied to a specific language. 
While the LLaMA2 tokenizer addresses unknown UTF-8 characters by converting them into bytes, this method significantly increases sequence length and hinders the speed of encoding and decoding Thai texts. 
Consequently, it presents difficulties for byte tokens and transformer encoders in adequately capturing the semantic nuances of Thai characters.
This allocation disparity necessitates additional vocab expansion, increasing the cost of adopting pretrained LLMs to low-to-medium resource languages.

Another important issue is the data transparency.
Although these open-source releases include the model's parameters and source code, some of the datasets used in training may not be clearly described or even inaccessible in some cases. 
This lack of transparency negatively affects the reproducibility of results and can cause unintentional data leakage during evaluation \cite{zhou2023don}.

To address the gaps in knowledge transferability and lack of data transparency, we introduce \emph{WangchanLion}, a Thai instruction-following model based on SEA-LION.
Our WangchanLion is based on SEA-LION, a pretrained generative text model specific to Southeast Asia languages developed by AI Singapore.
Our reasons for choosing SEA-LION as our base model are as follows.
First, the Thai vocab size of SEA-LION is 10,652 tokens, eliminating the need for vocab expansion when compared to LLaMA2.
Second, SEA-LION was pretrained on well-documented, open-source, publicly accessible datasets. 
%
WangchanLion continues the same spirit of data transparency initiated by SEA-LION, ensuring that our results are reproducible from scratch. 


Moreover, we perform experimental studies to compare WangchanLion with existing Thai-supporting instruction models. We chose to study machine reading comprehension (MRC) task since it is one of the most important components in question-answering applications.  We have found that WangchanLion outperforms two previous models with the same number of parameters in XQuAD \cite{artetxe-etal-2020-cross} and i\_app\_wiki\_qa\_squad benchmarks.

However, we have noticed some limitations of traditional MRC evaluation.
 To elaborate, measures like exact match accuracy and F1 score rely on token-level matching between predicted answer and reference answer, which can cause the following issues. 
First, the metrics unintentionally reward models that produce short answers. 
Second, answers that are semantically equivalent to reference answers but with different words are undesirably penalized. 
These properties can conflict with the objectives of using large language models for question answering in a conversational assistant manner, which we argue are correctness, helpfulness, conciseness, and contextuality.

To address the limitations of the current evaluation, we propose WangchanX-MRC-Eval, a human evaluation scheme for MRC. We found that WangchanLion tends to produce shorter answers, which results in less irrelevant and out-of-context information, but also provides less additional helpful contexts. These results help provide insights into how to further improve our LLMs for MRC in future works.  Furthermore, since human evaluation can cost a large amount of time and money, we also demonstrated using LLM to automate the evaluation process and improve the scalability of experiments.

Our contributions can be summarized as follows.
\begin{enumerate}
    \item We developed \emph{WangchanLion}, a Thai instruction-following language model that demonstrates an improved MRC performance over previous Thai-supported instruction models such as SeaLLMs and OpenThaiGPT.
    \item We propose \emph{WangchanX MRC Eval}, an automatic evaluation scheme in a human evaluation manner. This addresses the limitations of traditional MRC regarding the token-level matching between predicted answers and reference answers. 
    \item We publicly release the codes for WangchanX MRC Eval and pre-trained model weights of WangchanLion.
\end{enumerate}

\section{Instruction Tuning}


\subsection{Data}

To obtain WangchanLion, we perform instruction tuning using English and Thai datasets of 48,084,781 and 84,441,163 tokens, respectively.
The English instruction data consists of around 500,000 instruction pairs coming from the following sources.

\begin{itemize}
    \item unified-chip2, infild-bpedia, OpenAI Summarize TL;DR, and HC3 human from OIG-small-chip2, an Open Instruction Generalist dataset organized by LAION-AI~\cite{oig}.
    

    \item The entire collection of DataBricks Dolly, an instruction-following dataset from Databricks employees, including brainstorming, classification, closed QA, generation, information extraction, open QA, and summarization~\cite{DatabricksBlog2023DollyV2}.

    \item Dolphin, a FLANv2 dataset augmented with GPT-4 completions~\cite{dolphin}.

    \item Open-Platypus, an LLM logical reasoning dataset comprising PRM800K, openbookQA, SciBench, etc. We filter out Non-commercial datasets such as ReClor and ScienceQA~\cite{platypus2023}.
\end{itemize}

Our Thai datasets consist of a small portion of Thai instruction data. 
\begin{itemize}
    \item iapp\_wiki\_qa\_squad, an extractive question-answering dataset consisting of context, question and answer annotations from Thai Wikipedia articles \cite{kobkrit_viriyayudhakorn_2021_4539916}.
\end{itemize}
 \begin{itemize}
     \item Thaisum, a Thai text summarization dataset obtained from several online news websites \cite{chumpolsathien_2020}.
 \end{itemize}
\begin{itemize}
    \item XL-Sum, a dataset for abstractive summarization of professionally annotated article-summary pairs from BBC. The dataset covers 44 languages, including Thai \cite{hasan-etal-2021-xl}.
\end{itemize}

\begin{itemize}
    \item Han instruct v1.0 is an instruction following dataset organized by PyThaiNLP \cite{han_dataset}.
\end{itemize}

For English-Thai cross-lingual transfer, we used two cross-lingual sources.
\begin{itemize}
    \item xP3x (Crosslingual Public Pool of Prompts eXtended), a collection of prompts \& datasets across 277 languages and 16 NLP tasks, including Thai. We use only Thai~\cite{muennighoff2022crosslingual}.

    \item Scb\_mt\_enth\_2020, an English-Thai Parallel dataset for machine translation \cite{lowphansirikul2020scb}. We applied the MT dataset for improving the cross-lingual capability as demonstrated in previous LLM works~\cite{zhu2023extrapolating,cahyawijaya-etal-2023-instructalign,ranaldi2023empowering}. 
\end{itemize}



\subsection{Parameter-Efficient Finetuning (PEFT)}

Pre-trained generative models have the ability to predict the next token but still struggle to follow human instructions ~\cite{ouyang2022training}. To address this issue, we employ a strategy known as Parameter-Efficientr Finetuning (PEFT). PEFT utilizes sets of (Instruction, Input, Output) format tuples, a straightforward format prevalent in numerous NLP datasets. Following supervised fine-tuning, the model gains the ability to understand and follow human instructions effectively, as demonstrated by various instruction-tuned Language Models (LLMs) like InstructGPT~\cite{ouyang2022training} and Alpaca~\cite{alpaca}.

\begin{table}[htbp]
    \centering
    \begin{tabular}{ll} \hline
         Hyper-Parameter& Value\\ \hline
        QLoRa Rank & 512\\ \hline
        QLora Alpa & 512\\ \hline
        QLoRa Dropout & 0.05\\ \hline
        QLoRa Target Modules & down\_proj,out\_proj,up\_proj,Wqkv\\ \hline
        Epochs & 4 \\ \hline
        Learning Rate & 3e-4\\ \hline
        Batch Size & 2\\ \hline
 Optimizer& paged-adamw-32bit\\ \hline
        Floating Point Precision & bfoat16\\ \hline
    \end{tabular}
    \caption{Hyperparameter settings used for fine-tuning  WangchanLion.}
    \label{tab:hyper}
\end{table}


To enhance computational efficiency, we pack multiple sequences of training examples into a single sequence ~\cite{raffel2023exploring}. This implies that a batch can encompass one or more training examples, with each sample separated by an end-of-sequence (eos) token. We utilize parameter-efficient fine-tuning with QLoRa.\cite{dettmers2023qlora}. The hyperparameters used are listed in the table \ref{tab:hyper}. 



\section{Machine Reading Comprehension (MRC) Evaluation}

\begin{figure}[htbp]
    \centering
    \includegraphics[width=0.99\textwidth]{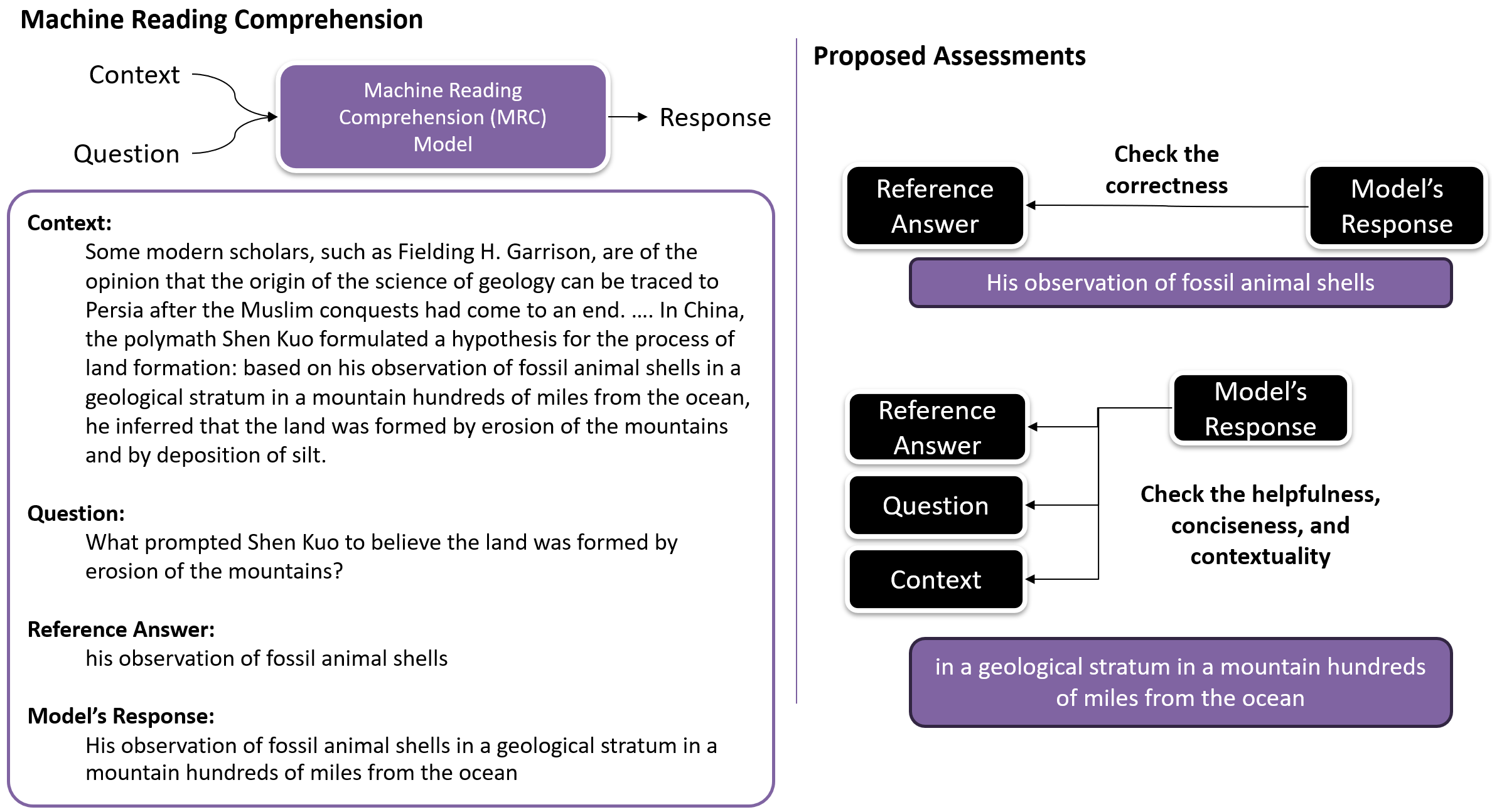}
    \caption{
    Proposed MRC Evaluation
    }
    \label{fig:MRCEVal}
\end{figure}

We evaluate the model from the final training checkpoint. We focus our evaluation on machine reading comprehension (MRC) since it is vital in a question-answering solution like retrieval-augmented generation (RAG)~\cite{gao2024retrievalaugmented}.  
As shown in Figure~\ref{fig:MRCEVal}, each MRC evaluation consists of four components: context, question, reference answer, and response from the model being assessed. 
Conventionally, an assessment is conducted by comparing the response with the reference answer to check the correctness.
In this study, we propose three additional assessments to check the response's helpfulness, conciseness, and contextuality. 

The evaluation consists of three studies: traditional extractive QA evaluation, human evaluation, and LLM-automated evaluation. 
\begin{description}
\item [Extractive QA Evaluation  (Section~\ref{sec:ExtractiveQAEval}):] 
To understand the strengths and limitations of the traditional MRC evaluation approach, we include an extractive QA evaluation benchmark, XQuAD \cite{artetxe-etal-2020-cross}.
We compare the models' responses against their respective reference answers using F1. 
A high score indicates the model's capability to generate responses faithful to the reference answers. 

\item [Human Evaluation  (Section~\ref{sec:HumanEval}):] 
We propose a more holistic approach to assessing responses from generative models.
Our evaluation method consists of four yes-no questions checking the correctness of the response and the qualities of additional information regarding its helpfulness, conciseness, and contextuality.

\item [LLM Evaluation  (Section~\ref{sec:LLMEval}).] While human evaluation provides a trusted gold standard, it has drawbacks in terms of cost and scale. 
We derive a more economical means to assess the correctness, helpfulness, conciseness, and contextuality in an automated fashion. 
In particular, we leverage a large language model like GPT-3.5 and GPT-4 to assess the responses. 
We compare their results to those obtained from the human evaluation for quality control.

\end{description}


\subsection{Traditional Extractive QA Evaluation} 
\label{sec:ExtractiveQAEval}

In this study, we use two datasets, XQuAD \cite{artetxe-etal-2020-cross} and i\_app\_wiki\_qa\_squad, to assess three of 7B models: OpenThaiGPT, SeaLLM V2, and WangchanLion. 
%
%
As presented in Figure~\ref{fig:MRCEVal}, each MRC evaluation is conducted by providing the context and question to the model and obtaining its response.  
This extractive evaluation only checks the correctness of the response by comparing it with the reference answer using F1.
A higher F1 score indicates a higher degree of text similarity concerning the reference answer.
Table~\ref{tab:f1score} displays results in two settings: 0-shot and 1-shot.
For 0-shot, the model received only the context and question to generate a response.
For 1-shot, the context and question were accompanied by an example tuple {\tt (context, question, reference answer)} as a few-shot in-context learning sample. 
We can see that WangchanLion obtained higher F1 scores than OpenThaiGPT and SeaLLM in both 0-shot and 1-shot settings for both datasets.

While extractive QA benchmarks are widely used in generative model evaluations due to their simplicity, it is important to understand their limitations.
First, reference answers in an extractive QA benchmark such as XQuAD tend to be short. 
As a result, they cannot measure the ability of a generative model to provide contextually rich answers.
Second, F1 can measure only text similarity.
As a result, a semantically correct answer can be penalized for using words different from the respective reference answer. 
These two limitations necessitate a more holistic way to assess generative MRC responses. 


\begin{table}[h]
    \centering
    \begin{tabular}{|c|c|c|c|c|} \hline 
         &  \multicolumn{2}{|c|}{XQuAD}&  \multicolumn{2}{|c|}{iapp\_wiki\_qa\_squad}\\ \hline  
         Model&  0-shot&  1-shot&  0-shot& 1-shot\\ \hline  
         OpenThaiGPT&  27.3487&  34.3104&  40.0614& 46.6883\\ \hline  
         SeaLLM V2&  16.1104&  25.7399&  23.6425& 28.9934\\ \hline 
         WangchanLion&  \textbf{45.8763}& \textbf{49.9145}&  \textbf{58.9051}& \textbf{62.9776}\\ \hline 
    \end{tabular}
   \\
    \caption{Experimental results on F1 score in 0-shot, 1-shot setting}
    \label{tab:f1score}
\end{table}


\subsection{Human Evaluation}
\label{sec:HumanEval}

We design a human evaluation scheme to overcome the limitations stated in the previous subsection. 
The main benefit of using human judgment is the flexibility to assess generative models according to human expectations.
However, we also want to keep things as objective as possible so that the evaluation can remain fair and consistent across different annotators and models. 
This approach also facilitates the automation of the evaluation process, allowing for scalability while maintaining consistent alignment with human expectations.

\textbf{Question Design.}
We design the assessment scheme through carefully structured yes-or-no questions to minimize subjective interpretation and ensure that the evaluation criteria are applied consistently across different annotators.
%
%
%
%
%
The evaluation consists of four questions assessing the response's correctness, helpfulness, conciseness, and contextuality. 
\begin{itemize}     
\item \textbf{Q1 - Correctness; The higher, the better [H]:}
\emph{The Answer is Correct concerning the Reference Answer. Do you agree or disagree?}
Determine if the given answer accurately matches the reference answer provided. The correctness here means the answer must directly correspond to the reference answer, ensuring factual accuracy.

\item \textbf{Q2 - Helpfulness; The higher, the better [H]:}
\emph{The Answer Includes Relevant, Additional Information from the Context. Do you agree or disagree?}
Assess whether the answer provides extra details that are not only correct but also relevant and enhance the understanding of the topic as per the information given in the context.

\item \textbf{Q3 - Irrelevancy; The lower, the better [L]:}
\emph{The Answer Includes Additional, Irrelevant Information from the Context. Do you agree or disagree?}
Check if the answer contains extra details that, while related to the context, do not directly pertain to the question asked. This information is not necessary to answer the question and is considered a digression.

\item \textbf{Q4 - Out-of-Context; The lower, the better [L]:}
\emph{The Answer Includes Information Not Found in the Context. Do you agree or disagree?}
Evaluate if the answer includes any information that is not included in the context. This information, even if correct, is extraneous as it goes beyond the provided text and may indicate conjecture or assumption.
\end{itemize}

\begin{figure}[htbp]
    \centering
    \includegraphics[width=0.45\textwidth]{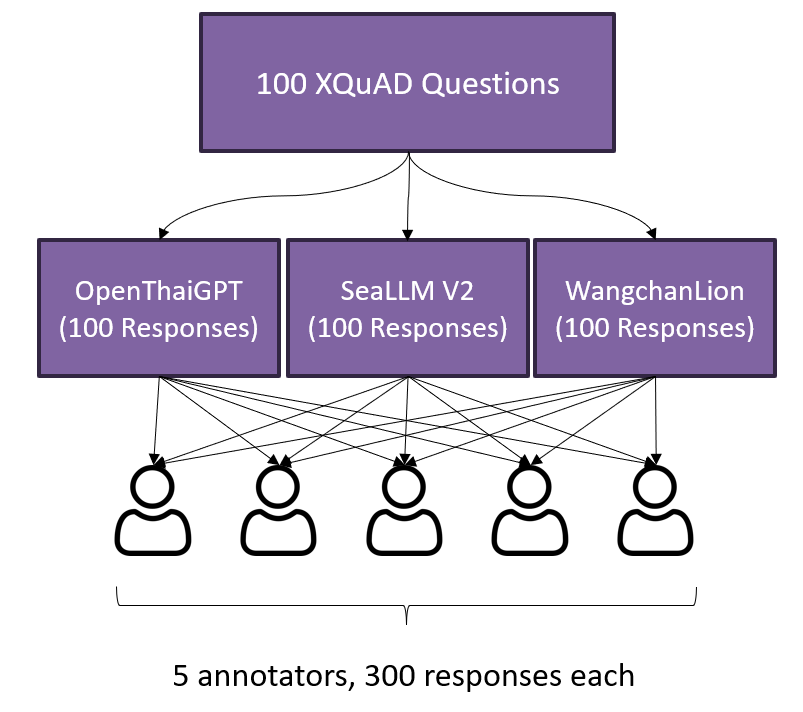}
    \caption{
    Human Response Collection Process
    }
    \label{fig:MRCEValCollection}
\end{figure}

\textbf{Human Response Collection.}
The human response annotation phase consists of three steps: training, screening, and deployment.
In the training step, candidates were given 15 sample responses with expected assessments to familiarize themselves with the task. Seven candidates participated in this step.
In the screening step, candidates were given 10 sample responses that they needed to answer. 
The training and screening samples were obtained from questions 1 to 100 from the XQuAD dataset. The responses is shown in Appendix \ref{appendix:question_human}.
In the deployment step, we selected candidates who scored more than 80\% as our annotators.
We obtained five annotators as a result. 
As shown in Figure~\ref{fig:MRCEValCollection},
these five annotators were assigned to assess responses from three models, OpenThaiGPT, SeaLLMs, and WangchanLion, answering 100 Questions in the XQuAD Dataset, bringing the total number of responses to 300.
\emph{Each annotator answered all four evaluation questions for all 300 responses}.

\textbf{Post Processing.}
After the annotation phase, the next step is to aggregate the responses from different annotators to reach a consensus. 
We employ a majority voting system to determine the final annotation for each question. 
For each of the four questions, the annotators' most common answer (either \emph{'agree'} or \emph{'disagree'}) became the final judgment for that item.

Table~\ref{tab:human_eval} displays the final results from the human evaluation.
We can see that despite the superior performance in terms of F1, WangchanLion is outperformed by SeaLLMs, judging based on Q1. 
In terms of additional information, SeaLLMs obtain higher Q2, Q3, and Q4 counts than WangchanLion.
These results suggest that SeaLLMs' lower F1 score is attributed to the tendency of the model to generate additional information instead of just giving a direct answer. 
A high Q2 count also shows that SeaLLms can identify helpful additional information within the provided context, a highly desirable feature in generative QA. 
However, this also comes with a risk of including irrelevant and out-of-context information, as shown in the results from Q3 and Q4.
The F1 and human-eval results imply that WangchanLion tends to generate results more akin to XQuAD-style extractive answers than the other two competitors.
SeaLLMs can produce additional information better aligned with human expectations than OpenThaiGPT regarding helpfulness and relevance while preserving contextuality. 
\begin{table}[htbp]
    \centering
    \begin{tabular}{|c|r|r|r|r|r|} \hline 
         Model&  Q1 &  Q2  &  Q3  & Q4 & Num Tokens\\ 
              &  Correctness &  Helpfulness &  Irrelevancy & Out-of-context &\\ 
              &  [H] & [H] & [L] & [L] & \\ \hline         
         OpenThaiGPT&  60&  38&  30& 32 & 10.35\\  \hline 
         SeaLLM V2&  \textbf{80}&  \textbf{80}&  \textbf{20}& 45 & 27.81\\  \hline 
         WangchanLion&  67&  19&  23& \textbf{5}& 5.50\\ \hline
    \end{tabular}
    \caption{Human evaluation scores from the four questions assessing the correctness of the answer (Q1) and the quality of additional information in terms of helpfulness (Q2), irrelevancy (Q3), and out-of-context information (Q4). [H]: The higher, the better. [L]: The lower, the better. Each figure in the Q1-Q4 columns is the count of responses, aggregated from majority votes from five annotators.  Num tokens denote the number of tokens obtained with PythaiNLP word tokenizer \cite{phatthiyaphaibun-etal-2023-pythainlp}.}
    \label{tab:human_eval}
\end{table}





\subsection{LLM-Automated Evaluation}
\label{sec:LLMEval}

To provide a more scalable means of evaluation, we derived a method to assess models' responses in an automated fashion.
We tested three options to automate our assessment: GPT-4 (gpt-4), GPT-3.5 (gpt-3.5-turbo), and Gemini (Gemini Pro).
As shown in Table~\ref{tab:LLM_comparison}, we found GPT-4 to be the most aligned with human evaluation results presented in the previous subsection and, hence, chosen as our assessor.  The prompt is shown in Appendix \ref{appendix:eval_prompt}.
%
\begin{table}[htbp]
\setlength{\tabcolsep}{3pt}
\renewcommand{\arraystretch}{1.3}
\centering
\begin{tabular}{|l|lll|lll|lll|lll|lll|}
\hline
\multirow{2}{*}{Assessor}
        & \multicolumn{3}{c|}{Q1: Correctness}                                           & \multicolumn{3}{c|}{Q2: Helpfulness}                                           & \multicolumn{3}{c|}{Q3: Irrelevancy}                                           & \multicolumn{3}{c|}{Q4: Out-of-Context}                                           & \multicolumn{3}{c|}{Overall}                                      \\ \cline{2-16}
        & \multicolumn{1}{c}{P} & \multicolumn{1}{c}{R} & \multicolumn{1}{c|}{F1}  & \multicolumn{1}{c}{P} & \multicolumn{1}{c}{R} & \multicolumn{1}{c|}{F1}  & \multicolumn{1}{c}{P} & \multicolumn{1}{c}{R} & \multicolumn{1}{c|}{F1}  & \multicolumn{1}{c}{P} & \multicolumn{1}{c}{R} & \multicolumn{1}{c|}{F1} & \multicolumn{1}{c}{P} & \multicolumn{1}{c}{R} & \multicolumn{1}{c|}{F1} \\ 
\hline
Gemini& 95.90& 90.34&    93.03& 89.80& 32.12&    47.31& 55.56& 13.70&    21.98& 61.11& 26.83&    37.29& \textbf{88.26}& 52.71&    66.00\\ \hline
GPT-3.5 & 91.08& 93.72&    92.38& 69.33& \textbf{75.91}&    \textbf{72.47}& \textbf{61.70}& 39.73&    48.33& 50.00& 43.90&    46.75& 75.31& \textbf{72.75}&    74.01\\ \hline
GPT-4   & \textbf{98.98}& \textbf{94.20}&    \textbf{96.53}& \textbf{94.29}& 48.18&    63.77& 55.17& \textbf{65.75}&    \textbf{60.00}& \textbf{75.41}& \textbf{56.10}&    \textbf{64.34}& 85.54& 71.14&    \textbf{77.68}\\ \hline
\end{tabular}
\caption{LLMs-automated evaluation compared with human evaluation. P R and F1 denote precision, recall, and F1 score computed by comparing the evaluation output of each LM compared to human responses.}
\label{tab:LLM_comparison}
\end{table}



Table~\ref{tab:gpt_eval} presents the results from the LLM-automated evaluation using GPT-4.
For ease of comparison, the three LLMs used in previous comparisons are displayed in the first three rows. 
Similar to the human evaluation results, SeaLLM is the best performer in terms of Q1 and Q2 among the three models.
For Q3 and Q4, the results also conform with the human evaluation.
WangchanLion obtained lower counts than the other two models, showing that the model has a lower risk of including irrelevant and out-of-context information.

The table also includes additional models to provide a broader picture: OpenThaiGPT 13B, LLaMA 7B, LLaMA 13B, PolyLM-Chat 13B and Typhoon-instruc-0130 ~\cite{pipatanakul2023typhoon} which is a new Thai instruction-following API developed by SCB 10X.
We can see that OpenThaiGPT 13B shows a slight improvement from OpenThaiGPT 7B.
typhoon-instruct-0130 are the best performers regarding Q1 and SeaLLM V2 are second regarding Q1, but they also have a high score in terms of Irrelevancy (Q3) and Out-of-context (Q4).
%
%
\begin{table}[htbp]
    \centering
    \begin{tabular}{|r|r|r|r|r|r|} \hline 
         Model&  Q1 &  Q2  &  Q3  & Q4 & Num Tokens\\ 
              &  Correctness &  Helpfulness &  Irrelevancy & Out-of-context &\\ 
              &  [H] & [H] & [L] & [L] & \\
              \hline 

         OpenThaiGPT 7B&  58&  14&  29& 28&10.35\\ \hline 
         SeaLLM V2&  75&  \textbf{46}&  32& 30&27.81\\ \hline 
         WangchanLion&  64&  10&  26& \textbf{3}&5.50\\ \hline \hline
         OpenThaiGPT 13B&  59&  26&  37& 34&17.08\\ \hline 
         PolyLM-Chat 13B&  73& 17& \textbf{16}& 4&11.96\\\hline
         Typhoon-instruct-0130& \textbf{76}& 28& 24& 22&18.33\\\hline
    \end{tabular}
    \vspace{2mm}
    \caption{GPT evaluation scores from the four questions assessing the correctness of the answer (Q1) and the quality of additional information in terms of helpfulness (Q2), irrelevancy (Q3), and out-of-context information (Q4). [H]: ``The higher, the better. [L]: ``The lower, the better. For the assessment cost using GPT-4, as of March 2024, assessing one model with 100 questions incurred approximately \$4.00. The assessment comprising seven models was completed with less than \$30.00 on the OpenAI API. }
    \label{tab:gpt_eval}
\end{table}

\subsection{Human Prefer Short or Long answers}
\begin{figure}[htbp]
    \centering
    \includegraphics[width=0.99\textwidth]{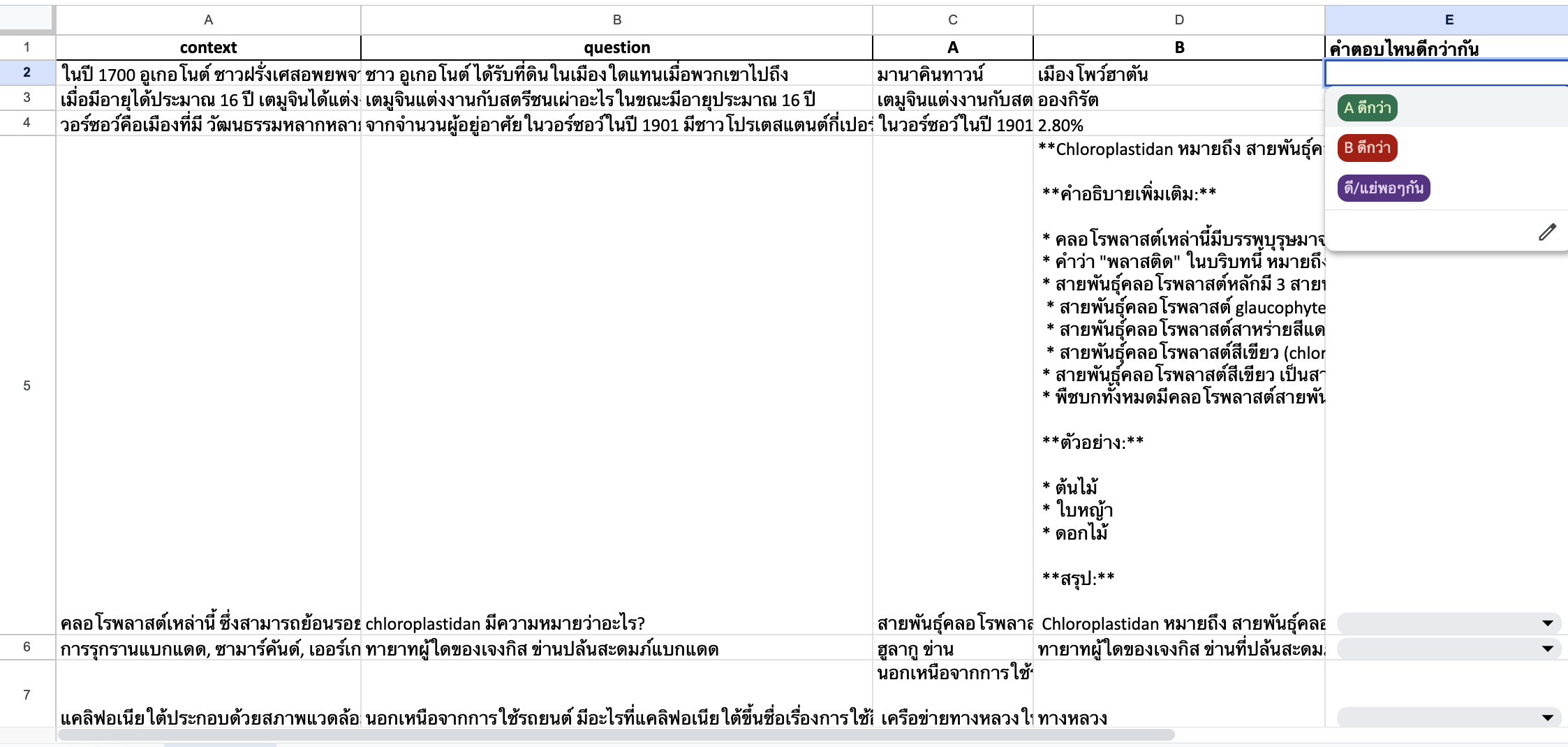}
    \caption{
    User interface of the human response collection.
    }
    \label{fig:Vicuna}
\end{figure}
Based on our evaluation criteria, we have established that users prefer longer answers over short ones. To ensure user satisfaction, we utilize Vicuna evaluation~\cite{zheng2023judging} in 50 questions by submitting our evaluation to 10 humans. We rely on XQuAD answers for quick answers, while for detailed and comprehensive responses, we use the text generated from the model. Our Data collection uses Google Sheets as shown in Figure \ref{fig:Vicuna} and asks humans: \foreignlanguage{thai}{คำตอบไหนดีกว่ากัน} (Which answer is better?). The answers are A \foreignlanguage{thai}{ ดีกว่า} (A is better), B \foreignlanguage{thai}{ ดีกว่า} (B is better), and \foreignlanguage{thai}{ ดี/แย่พอๆกัน} (Tie).

After we got the result, 31 evaluations resulted in long answers, 12 in short answers, and 7 in a tie. We found humans prefer longer answers over short ones. The results show humans want long answers more than quick ones, and our evaluation criteria can evaluate the long answers to create the answer humans wish.

\section{Conclusion and Future work}
We release WangChanLion, an open-source Thai instruction following model focusing on Question Answering. We also proposed a new method for model evaluation by GPT4 in the MRC task. 
For evaluation, we want high Q1 and Q2 counts while keeping those of Q3 and Q4 low. 
This implies that our model should produce responses that are not only correct but also accompanied by additional information that satisfies the criterion of Q2.
However, optimizing for the Q2 count may come with the risk of extraneous information reflected in Q3 and Q4 count differentials between WangchanLion and the other two methods.
The main challenge here is introducing more useful information while keeping extraneous information in check. 

In our upcoming research, we plan to delve deeper into the exploration of PEFT and explore alternative alignment approaches like Direct Preference Optimization (DPO) ~\cite{rafailov2023direct} . Additionally, we aim to focus on selecting datasets of superior quality.


\bibliographystyle{unsrt}  
\bibliography{references}
\newpage
\appendix
\renewcommand*{\thesection}{\Alph{section}}

\section{Model prompt for XQuAD}
\label{appendix:model_prompt}
\subsection{WangchanLion}

\begin{framed}
You are a helpful assistant. Read the context and answer the question.

\foreignlanguage{thai}{ พื้นหลัง}:

\{Context\}

\foreignlanguage{thai}{ คำถาม}:

\{question\}

\foreignlanguage{thai}{ ตอบ}:
\end{framed}

\subsection{OpenThaiGPT}
\begin{framed}
Below is an instruction that describes a task. Write a response that appropriately completes the request. Read the context and answer the question.

\#\#\# Instruction:

Context (\foreignlanguage{thai}{บริบท}): \{Context\}

Question (\foreignlanguage{thai}{คำถาม}): \{Question\}

\#\#\# Response:
\end{framed}
\subsection{SeaLLM}
\begin{framed}
<|im\_start|>system You are a helpful assistant. Read the context and answer the question.</s><|im\_start|>user

Context (\foreignlanguage{thai}{บริบท}): 

\{Context\}

Question (\foreignlanguage{thai}{คำถาม}):

\{Question\}</s><|im\_start|>user
\end{framed}
\subsection{PolyLM}
\begin{framed}
You are a helpful assistant. Read the context and answer the question.

<|user|>

Context: \{Context\}

Question: \{Question\}

<|assistant|>
\end{framed}
\subsection{Typhoon-instruct-0130}
\begin{framed}
Below is an instruction that describes a task. Write a response that appropriately completes the request.

\#\#\# Instruction: 

Read the context and answer the question

\#\#\# Context:

\{Context\}

\#\#\# Question:

\{Question\}
\end{framed}
\section{Question for expected assessments}
\label{appendix:question_human}
\subsection{Q15}

In the training step, we were give some example from 15 sample responses with expected assessments to familiarize themselves with the task. All 15 sample responses for training step can see at \url{https://github.com/vistec-AI/WangchanLion/tree/main/mrc_eval/human_data_collection}.

Examples:

\begin{enumerate}
    \item \foreignlanguage{thai}{ ข้อความบริบท}: \foreignlanguage{thai}{ หลังช่วงยุคปี 1940 สไตล์กอธิคในวิทยาลัยเริ่มเปิดทางให้กับสไตล์ทันสมัย ในปี 1955 Eero Saarinen ได้ทำสัญญาเพื่อทำแผนแม่บทฉบับที่สอง ซึ่งนำไปสู่การสร้างอาคารทั้งทางเหนือและใต้ของ Midway ประกอบด้วย Laird Bell Law Quadrangle (อาคารที่ถูกออกแบบโดย Saarinen) ชุดอาคารศิลปะ อาคารที่ถูกออกแบบโดย ลูทวิช มีส ฟัน แดร์ โรเออ ให้กับ โรงเรียนบริหารงานสังคมสงเคราะห์ ของมหาวิทยาลัย อาคารที่ซึ่งกลายเป็นบ้านของ โรงเรียนนโยบายสาธารณะแฮร์ริส โดยเอ็ดเวิร์ด ดอเรลล์ สโตน และห้องสมุด Regenstein อาคารที่ใหญ่ที่สุดในมหาวิทยาลัย สิ่งก่อสร้างบรูทัลลิสต์ออกแบบโดยวอลเตอร์ เน็ตช์แห่งบริษัท Skidmore, Owings \& Merrill แผนแม่แบบอื่น ๆ ถูกออกแบบในปี 1999 และอัพเดทในปี 2004 สร้างศูนย์กรีฑา Gerald Ratner (2003) Max Palevsky Residential Commons (2001) หอพักนักศึกษาทิศใต้และโรงอาหารส่วนกลาง (2009) โรงพยาบาลเด็กแห่งใหม่ และสิ่งก่อสร้าง, ส่วนขยาย และการบูรณะอื่น ๆ ในปี 2011 มหาวิทยาลัยสร้างห้องสมุดทรงโดมแก้ว Joe and Rika Mansueto ซึ่งมอบห้องอ่านหนังสือขนาดใหญ่สำหรับห้องสมุดมหาวิทยาลัยและป้องกันความจำเป็สในการฝากหนังสือนอกมหาวิทยาลัย}

\foreignlanguage{thai}{ คำถาม}: \foreignlanguage{thai}{ โรงเรียนนโยบายสาธารณะใดที่ได้อยู่ในอาคารที่ ลูทวิช มีส ฟัน แดร์ โรเออ ออกแบบ?}

\foreignlanguage{thai}{ คำตอบอ้างอิง}: \foreignlanguage{thai}{ โรงเรียนนโยบายสาธารณะแฮร์ริส}

\foreignlanguage{thai}{ คำตอบจากโมเดล}: \foreignlanguage{thai}{ แฮร์ริส}

Q1: \foreignlanguage{thai}{ ไม่เห็นด้วย}

Q2: \foreignlanguage{thai}{ ไม่เห็นด้วย}

Q3: \foreignlanguage{thai}{ ไม่เห็นด้วย}

Q4: \foreignlanguage{thai}{ ไม่เห็นด้วย}
\item \foreignlanguage{thai}{ ข้อความบริบท}: \foreignlanguage{thai}{ กวิชาการสมัยใหม่บางคน เช่น ฟีลดิง เอช. การ์ริสัน เชื่อว่าเราสามารถสาวต้นกำเนิดของศาสตร์แห่งธรณีวิทยากลับไปถึงดินแดน เปอร์เซีย หลังจากสิ้นสุดการพิชิตดินแดนโดยมุสลิม อาบู อัลเรฮัน อัลบิรูนี (973–1048 สากลศักราช) คือหนึ่งในนักธรณีวิทยาชาวเปอร์เซียคนแรกๆ ผลงานของเขารวมถึงการเขียนถึงธรณีวิทยาของอินเดีย โดยสันนิษฐานว่าครั้งหนึ่งอนุทวีปอินเดียเคยเป็นทะเลมาก่อน อิบิน ซีน่า (อาวิเซนน่า, ค.ศ. 981–1037) นักวิชาการชาวเปอร์เซีย นำข้อมูลจากวรรณกรรมเชิงวิทยาศาสตร์ของกรีซและอินเดียซึ่งไม่โดนทำลายโดยการพิชิตดินแดนโดยมุสลิมมานำเสนอคำอธิบายโดยละเอียดเกี่ยวกับการก่อตัวของภูเขา ต้นกำเนิดของแผ่นดินไหว และหัวข้ออื่นๆ ซึ่งใกล้เคียงกับธรณีวิทยาสมัยใหม่ ซึ่งเป็นรากฐานที่สำคัญของการพัฒนาทางวิทยาศาสตร์ในภายหลัง ในประเทศจีน พหูสูตคนหนึ่งชื่อว่า เซิ่น โค้ว (ค.ศ.1031–1095) ได้ตั้งสมมติฐานเกี่ยวกับการก่อตัวของพื้นดิน โดยอ้างอิงจาก การสังเกตฟอสซิลของเปลือกและกระดองสัตว์ ต่างๆ ในชั้นดินบนภูเขาซึ่งอยู่ห่างจากมหาสมุทรหลายร้อยไมล์ เขาอนุมานว่าพื้นดินเกิดจากการกัดกร่อนของภูเขาและการสะสมของตะกอน}

\foreignlanguage{thai}{ คำถาม}: \foreignlanguage{thai}{ อะไรทำให้ เซิ่น โค้ว เชื่อว่าพื้นดินเกิดจากการกัดกร่อนของภูเขา}

\foreignlanguage{thai}{ คำตอบอ้างอิง}: \foreignlanguage{thai}{ การสังเกตฟอสซิลของเปลือกและกระดองสัตว์ ต่างๆ}

\foreignlanguage{thai}{ คำตอบจากโมเดล}: \foreignlanguage{thai}{ เซิ่น โค้ว เชื่อว่าพื้นดินเกิดจากการกัดกร่อนของภูเขาและสะสมของตะกอน เขาเชื่อว่าภูเขาก่อตัวขึ้นบนแผ่นเปลือกโลกและค่อยๆ กัดกร่อนไปตามกาลเวลา เมื่อแผ่นเปลือกโลกเคลื่อนตัวและภูเขาไฟระเบิดขึ้น แผ่นดินก็เกิดขึ้นจากการสะสมของตะกอน}

Q1: \foreignlanguage{thai}{ ไม่เห็นด้วย}

Q2: \foreignlanguage{thai}{ ไม่เห็นด้วย}

Q3: \foreignlanguage{thai}{ เห็นด้วย}

Q4: \foreignlanguage{thai}{ เห็นด้วย}
\end{enumerate}

\subsection{Q10}

In the screening step, candidates were given 10 sample responses that
they needed to answer. This section give some example. All 10 sample responses for screening step can see at \url{https://github.com/vistec-AI/WangchanLion/tree/main/mrc_eval/human_data_collection}.

Examples:

\begin{enumerate}
    \item \foreignlanguage{thai}{ ข้อความบริบท}: \foreignlanguage{thai}{ ในทางตรงกันข้าม ระหว่างห้วงเวลาที่ตื่นอยู่เซลล์แสดงผลที่แตกต่างกัน อย่างเช่น เซลล์นักฆ่าตามธรรมชาติด้วยพิษ และ CTL (ไซโตท็อกสิก ที ลิมโฟไซต์) จะมีจำนวนพุ่งสูงขึ้นเพื่อกระตุ้นการตอบสนองที่มีประสิทธิภาพต่อจุลชีพก่อโรคใด ๆ ที่บุกรุกเข้ามา เช่นเดียวกัน ในระหว่างห้วงเวลาที่ตื่นอยู่ โมเลกุลต้านการอักเสบ อย่างเช่น คอร์ติซอล และแคทีโคลามีนส์ จะพุ่งสูง มีอยู่สองทฤษฎีที่อธิบายว่าเพราะเหตุใดสภาวะการอักเสบจึงได้รับการรักษาไว้สำหรับห้วงเวลาหลับ ทฤษฎีแรกคือ การอักเสบจะก่อให้เกิดความบกพร่องทางสติปัญญาและร่างกายอย่างรุนแรงหากเกิดขึ้นในระหว่างเวลาตื่น ทฤษฎีที่สองคือ การอักเสบอาจเกิดขึ้นในระหว่างการนอนหลับเนื่องจากในเวลานั้นมี เมลาโทนิน การอักเสบก่อให้เกิดภาวะเครียดที่เกิดจากออกซิเดชันเป็นอย่างมาก และการมีเมลาโทนินในระหว่างหลับสามารถต่อต้าน การผลิตอนุมูลอิสระ ได้เป็นอย่างดี}

\foreignlanguage{thai}{ คำถาม}: \foreignlanguage{thai}{ การอักเสบเกิดขึ้นในระหว่างเวลานอนหลับเนื่องจากการมีโมเลกุลอะไร}

\foreignlanguage{thai}{ คำตอบอ้างอิง}: \foreignlanguage{thai}{ เมลาโทนิน}

\foreignlanguage{thai}{ คำตอบจากโมเดล}: \foreignlanguage{thai}{ การอักเสบเกิดขึ้นในระหว่างเวลานอนหลับเนื่องจากการมีโมเลกุลที่เรียกว่าเมลาโทนิน (Melatonin) ซึ่งเป็นสารสื่อประสาทที่เกิดขึ้นในสมองและร่างกายของมนุษย์ในช่วงเวลานอนหลับ}

Q1: \foreignlanguage{thai}{ เห็นด้วย/ไม่เห็นด้วย}

Q2: \foreignlanguage{thai}{ เห็นด้วย/ไม่เห็นด้วย}

Q3: \foreignlanguage{thai}{ เห็นด้วย/ไม่เห็นด้วย}

Q4: \foreignlanguage{thai}{ เห็นด้วย/ไม่เห็นด้วย}
    \item \foreignlanguage{thai}{ ข้อความบริบท}: \foreignlanguage{thai}{ การศึกษาวิจัยยังพบด้วยว่ามีเชื้อโรคที่เกี่ยวข้องแต่ไม่เป็นที่รู้จักอีกสองชนิดซึ่งเป็นเคลด (สายพันธุ์ย่อย) ของจีโนมวาย.เพสติส ที่มีความเกี่ยวข้องกับการเสียชีวิตของผู้คนจำนวนมหาศาลในยุคกลาง มีการค้นพบว่าเชื้อโรคเหล่านี้ (ซึ่งเชื่อกันว่าสูญพันธุ์ไปแล้ว) เป็นบรรพบุรุษของสายพันธุ์วาย.เพสทิสในสมัยใหม่ คือ วาย.พี. โอเรียนทาลิส และ วาย.พี เมดิอีวาลิส และเชื่อกันว่า กาฬโรคอาจเข้าสู่ทวีปยุโรปสองระลอก การสำรวจหลุมศพของผู้เสียชีวิตจากกาฬโรคในฝรั่งเศสและอังกฤษแสดงให้เห็นว่าเชื้อกาฬโรคแบบแรกเข้าสู่ยุโรป ผ่านทางท่าเรือมาร์กเซย์ในราวเดือนพฤศจิกายน ปี 1347 และระบาดไปทั่วยุโรปภายในสองปีต่อมา และในที่สุดก็ไปถึงอังกฤษในฤดูใบไม้ผลิปี 1349 ซึ่งที่อังกฤษนี้โรคได้แพร่กระจายเป็นโรคระบาดสามโรคด้วยกัน การสำรวจหลุมศพของผู้เสียชีวิตจากกาฬโรคที่เมืองดัตช์แห่งแบร์เกน อป ซูม แสดงให้เห็นจีโนไทป์ของวาย.เพสทิสซึ่งเป็นสาเหตุของโรคระบาดซึ่งแพร่กระจายไปทั่วประเทศเนเธอร์แลนด์ ตั้งแต่ปี 1350 นั้นต่างกับที่ค้นพบในอังกฤษและฝรั่งเศส ซึ่งแสดงให้เห็นว่าการระบาดที่แบร์เกน อป ซูม (และเป็นไปได้ว่าในบริเวณอื่นๆ ของเนเธอร์แลนด์ตอนใต้) ไม่ได้มาจากการระบาดในอังกฤษหรือฝรั่งเศสในปี 1349 โดยตรง และยังชี้ให้เห็นด้วยว่าโรคระบาดระลอกที่สองต่างกับที่เกิดขึ้นในอังกฤษและฝรั่งเศส โดยอาจเข้ามาสู่เนเธอร์แลนด์ผ่านทางนอร์เวย์ เมืองในแถบฮันเซียติก หรือบริเวณอื่น}

\foreignlanguage{thai}{ คำถาม}: \foreignlanguage{thai}{ วาย.เพสทิสระบาดไปถึงอังกฤษเมื่อใด}

\foreignlanguage{thai}{ คำตอบอ้างอิง}: \foreignlanguage{thai}{ 1349}

\foreignlanguage{thai}{ คำตอบจากโมเดล}: \foreignlanguage{thai}{ ฤดูใบไม้ผลิ 1349}

Q1: \foreignlanguage{thai}{ เห็นด้วย/ไม่เห็นด้วย}

Q2: \foreignlanguage{thai}{ เห็นด้วย/ไม่เห็นด้วย}

Q3: \foreignlanguage{thai}{ เห็นด้วย/ไม่เห็นด้วย}

Q4: \foreignlanguage{thai}{ เห็นด้วย/ไม่เห็นด้วย}
\end{enumerate}

\section{Config}
\label{appendix:config}
\subsection{GPT-4}

Base openai client v0.28.

\begin{itemize}
    \item temperature: 0.2
    \item max\_tokens: 1024
\end{itemize}

\subsection{GPT-3.5}

Base openai client v0.28.

\begin{itemize}
    \item temperature: 0.2
    \item max\_tokens: 1024
\end{itemize}

\subsection{Gemini-Pro}
\begin{itemize}
    \item temperature: 0.9
    \item top\_p: 1
    \item top\_k: 1
    \item max\_tokens: 2048
\end{itemize}

\section{Evaluation Prompt}
\label{appendix:eval_prompt}
We use some prompts like human questions with the fill "This is very important to my career" \cite{li2023large} that boost our performance.

\subsection{GPT-4/GPT-3.5 Evaluation Prompt}

\textbf{system\_prompt}
\begin{framed}
Please evaluate these answers based on their accuracy and relevance to the provided passage that based on the Criteria:

1. The Answer is Correct concerning the Reference Answer. Do you agree or disagree?

Determine if the given answer accurately matches the reference answer provided. The correctness here means the answer must directly correspond to the reference answer, ensuring factual accuracy.

2. The Answer Includes Relevant, Additional Information from the Context. Do you agree or disagree?

Assess whether the answer provides extra details that are not only correct but also relevant and enhance the understanding of the topic as per the information given in the context.

3. The Answer Includes Additional, Irrelevant Information from the Context. Do you agree or disagree?

Check if the answer contains extra details that, while related to the context, do not directly pertain to the question asked. This information is not necessary for answering the question and is considered a digression.

4. The Answer Includes Information Not Found in the Context. Do you agree or disagree?

Evaluate if the answer includes any information that is not included in the context. This information, even if correct, is extraneous as it goes beyond the provided text and may indicate conjecture or assumption.

This is very important to my career.
\end{framed}
\textbf{user\_prompt}
\begin{framed}
Passage: \{context\}

Question: \{question\}

Reference Answer: "\{reference\_answer\}"

Prediction Answer: "\{prediction\_answer\}"
\end{framed}

\subsection{Gemini-Pro  Evaluation Prompt}

\textbf{prompt}
\begin{framed}
Please evaluate these answers based on their accuracy and relevance to the provided passage that based on the Criteria:

1. The Answer is Correct concerning the Reference Answer. Do you agree or disagree?

Determine if the given answer accurately matches the reference answer provided. The correctness here means the answer must directly correspond to the reference answer, ensuring factual accuracy.

2. The Answer Includes Relevant, Additional Information from the Context. Do you agree or disagree?

Assess whether the answer provides extra details that are not only correct but also relevant and enhance the understanding of the topic as per the information given in the context.

3. The Answer Includes Additional, Irrelevant Information from the Context. Do you agree or disagree?

Check if the answer contains extra details that, while related to the context, do not directly pertain to the question asked. This information is not necessary for answering the question and is considered a digression.

4. The Answer Includes Information Not Found in the Context. Do you agree or disagree?

Evaluate if the answer includes any information that is not included in the context. This information, even if correct, is extraneous as it goes beyond the provided text and may indicate conjecture or assumption.

This is very important to my career.

Passage: \{context\}

Question: \{question\}

Reference Answer: "\{reference\_answer\}"

Prediction Answer: "\{prediction\_answer\}"
\end{framed}
\end{document}